\documentclass[conference,a4paper]{APSIPA2012}
\usepackage{multirow}
\usepackage{amsmath}
\usepackage[psamsfonts]{amssymb}
\usepackage{amsxtra}
\usepackage{threeparttable}
\usepackage{epsfig}
\usepackage{epstopdf,graphicx}
\usepackage{color}
\usepackage{epstopdf}
\usepackage{epsfig}
\usepackage{url}

\begin{document}

\title{Transfer Learning for Speech and Language Processing}

\author{%
\authorblockN{%
Dong Wang and Thomas Fang Zheng
}
\authorblockA{%
1. Center for Speech and Language Technologies (CSLT)  \\
Research Institute of Information Technology, Tsinghua University \\
2. Tsinghua National Lab for Information Science and Technology \\
Beijing, 100084, P.R.China
}
}

\maketitle
\thispagestyle{empty}

\begin{abstract}

Transfer learning is a vital technique that generalizes models trained for one setting or task to other settings or tasks.  For example in speech recognition, an acoustic model trained for one language can be used to recognize speech in another language, with little or no re-training data. Transfer learning is closely related to multi-task learning (cross-lingual vs. multilingual), and is traditionally studied in the name of `model adaptation'. Recent advance in deep learning shows that transfer learning  becomes much easier and more effective with high-level abstract features learned by deep models, and the `transfer' can be conducted not only between data distributions and data types, but also between model structures (e.g., shallow nets and deep nets) or even model types  (e.g., Bayesian models and neural models). This review paper summarizes some recent prominent research towards this direction, particularly for speech and language processing. We also report some results from our group and highlight the potential of this very interesting research field\footnote{This survey will be continuously updated online () to reflect the recent progress on transfer learning. }.

\end{abstract}

\section{Introduction}
\label{sec:intro}

Machine learning (ML) techniques have been extensively exploited in modern speech and language processing research~\cite{martin2000speech,benesty2008springer,deng2013machine}. Among the rich family of ML models and algorithms, transfer learning is among the most interesting. Generally speaking, transfer learning involves all methods that utilize any auxiliary resources (data, model, labels, etc.) to enhance model learning for the target task~\cite{pan2010survey,taylor2009transfer,bengio2012deep,lu2015transfer}. This is very important for speech and language research, since human speech and languages are so diverse and imbalanced. There are more than $5,000$ languages around the world, and the number is even bigger if dialects are counted. Among this big family, $389$ languages (nearly 6\%) account for 94\% of the word's population, and the rest thousands languages are spoken by very few people.\footnote{https://www.ethnologue.com/statistics} Even for the $389$ `big' languages, only very few possess adequate resources (speech signal, text corpus, lexicon, phonetic/syntactic regulations, etc.) for speech and language research. If we talk about `rich-resource' languages, perhaps only English is in that category.
Additionally, resources in different domains are also highly imbalanced, even for English. This means that almost all research in speech and language confront the challenge of data sparsity. More seriously, human language is such dynamic that new words and domains emerge every day, and so no models learned at a particular time will remain valid forever.

With such diversity, variation, imbalance and dynamics, it is almost impossible for speech and language researchers to learn a model from one single data resource and then put it on the shelf. We have to resort to some more smart algorithms that can learn from multiple languages, multiple data, multiple domains and keep the model adapted. On the other hand, it would not be very controversial to argue that human speech and languages hold some common statistical patterns at both the signal and symbolic levels, so that learning from multiple resources is possible.

In fact, transfer learning has been studied for a long time in a multitude of research fields in speech and language processing, e.g., speaker adaptation and multilingual modeling in speech recognition, cross-language document classification and sentiment analysis. Most of the studies, however, are task-driven in their own research fields and seldom hold deep understanding about the position of their research in the whole picture of transfer learning. This prevents researchers from answering some important questions: how and in which conditions their methods work, what are possible alternatives of their methods, and what advantages can be achieved with different alternatives? In this paper, we will give a brief summary of the most promising transfer learning methods, particularly within the modern deep learning paradigm. Special focus will be put on the application of transfer learning in speech and language processing, and some recent results from our research team will be presented.

We highlight that it is not our goal to present an entire list of the transfer learning methods in this paper. Instead, the focus is put on the most promising approaches for speech and language processing. Even with such a constraint, the work on transfer learning is still too much to be enumerated, and we can only touch a small part of the plenty techniques. We decide to focus on two specific domains: speech recognition and document classification, particularly the most recent advances based on deep learning which is most relevant to our research. For more detailed surveys on transfer learning in broad research fields, readers are referred to the nice review articles from Pan, Taylor, Bengio and Lu~\cite{pan2010survey,taylor2009transfer,bengio2012deep,lu2015transfer} and the references therein.

The paper is organized as follows: Section~\ref{sec:tl} gives a quick review of the transfer learning approach, and Section~\ref{sec:speech} and Section~\ref{sec:lang} discuss application of transfer learning in speech processing and language processing respectively. The paper is concluded in Section~\ref{sec:con}, with some discussions for the future research directions in this very promising field.

\section{Transfer learning: A quick review}
\label{sec:tl}

The motivation of transfer learning can be found in the idea of "Learning to Learn", which stats that learning from scratch (tabula rasa learning) is often limited, and so past experience should be used as much as possible~\cite{thrun2012learning}. For instance, once we learned that a hard apple is often sour, this experience can be used when we select pears: we conjecture that hard pears are also sour. This idea and associated research trace back to $20$ years ago and were summarized in the NIPS 95 workshop on `Learning to Learn: Knowledge Consolidation and Transfer in Inductive Systems'~\cite{nips95}. Many ideas and research goals raised in that workshop last two decades and influence our research till today, though the data, models, algorithms, computing power have dramatically changed. Some of the recent developments were discussed in several workshops, e.g., the ICML 2011 workshop on unsupervised and transfer learning\footnote{http://clopinet.com/isabelle/Projects/ICML2011/}; the NIPS 2013 workshop on new directions in transfer and multitask\footnote{https://sites.google.com/site/learningacross/}‎; the ICDM 2015 workshop on practical transfer learning\footnote{https://sites.google.com/site/icdmwptl2015/home}. In this section, we review some of the most prominent approaches to transfer learning, particularly those have been applied to or are potential for speech and language processing.

\subsection{Categories of transfer learning}
\label{sec:tl:cat}

The initial idea of transfer learning is to reuse the experience/knowledge obtained already to enhance learning for new things. Depending on the relation of the `old things' (source) that we have learned and the `new things' (target) that we want to learn, a large amount of methods have been devised, in different names by different authors. A short list of these names include multitask learning, lifelong learning, knowledge transfer, knowledge consolidation, model adaptation, concept drift, covariance shift, etc. Different researchers hold different views for the categorization of these methods. For example, Pan and Yang~\cite{pan2010survey} believed transfer learning should really `transfer' something so multitask learning should be regarded as a different approach, while Bengio~\cite{bengio2012deep} treated transfer learning and multitask learning as synonyms.

In our opinion, the different learning methods mentioned above can be regarded as particular implementations of transfer learning applied in different conditions or by different ways. For example, model adaptation is applied to conditions where the data distributions of the source and target domains are clearly different, while covariance drift is applied to conditions where the distribution changes gradually. As another example, knowledge transfer is applied to the condition where the source model and target model are trained sequentially, while multi-task learning is applied to the condition where the source and target models are trained simultaneously. No matter what forms and properties the learning methods hold, what they all have in common is `the attempt to transfer knowledge from other sources to benefit the current inductive task', and the benefit of the transfer involves faster convergence, more robust models and less data sensitivity.

\begin{figure}[htb]
  \centering{
  \epsfig{file=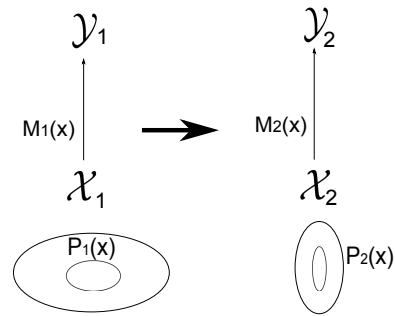, width=50mm}
  \caption{Relation of the conditional factors in the transfer learning paradigm. $\mathcal{X}_1$ and $\mathcal{Y}_1$ are the feature and label spaces respectively for the learning task in the source domain, and $\mathcal{X}_2$ and $\mathcal{Y}_2$ are for the learning task in the target domain. $M_1(x)$ and $M_2(x)$ represent the models in the source and target domains, respectively.}
  \label{fig:rel}
  }
\end{figure}

We can thus categorize transfer learning into several classes according to the conditions that they apply to. Following the taxonomy in~\cite{pan2010survey}, we use \emph{data} and \emph{task} as two conditional factors of transfer learning. For the data condition, it involves the feature space $\mathcal{X}$ (e.g., audio or text) and the distribution $P(X)$ of the feature (e.g., financial news and scientific papers); for the task condition, it involves the label space $\mathcal{Y}$ (e.g., speech phones or speaker identity) and the model $M(x)$ (e.g., probabilistic models or neural models). Any of the two components of the two conditional factors can be the same or different for the learning in the source and target domains, and their relation is shown in Fig.~\ref{fig:rel}. Note that if the feature space is different for the source and target domains, then their distributions are certainly different. Similarly, if the labels are different, then the models are regarded as different, although models from the same family might be used in the source and target domains.

According to whether the conditional factors (data and task) of the learning in the source and target domains are different or not, transfer learning methods can be categorized into several classes. Table~\ref{tab:transf} shows some of the most popular transfer learning approaches that are applicable in different conditions. In the table, `+' means the corresponding conditional factor is the same for the source and target domains, while `-' means different. Note that transfer learning is such a large research field and it is impossible to classify all the methods in such a simple way. For example, an important factor that discriminates different learning methods is whether or not the data in the source and target domains are labelled, which is not clearly reflected in the table (though we will discuss the related issue in the next section). Anyway, Table~\ref{tab:transf} gives a rough picture how big the family of transfer learning methods and how they can be categorized according to the conditional factors.

\begin{table*}[htb!]
\caption{Categories of transfer learning}
\label{tab:transf}
\center{
\begin{tabular}{l|c|c|c|c|c}
  \hline
     \multicolumn{2}{c|}{}  &\multicolumn{2}{c|}{$\mathcal{Y}+$} & \multicolumn{2}{c}{$\mathcal{Y}-$} \\
  \hline
   \multicolumn{2}{c|}{}      &M(x)+ &  M(x) - & \multicolumn{2}{c}{} \\
  \hline
  \hline
  $\mathcal{X}+$ & P(X)+ & Conventional ML & Model transfer\cite{hinton2014distilling} &\multicolumn{2}{c}{Multitask learning\cite{caruana1997multitask}}\\
  \cline{2-6}
                 & P(X)- & Model Adaptation\cite{gauvain1994maximum,leggetter1995maximum}, incremental learning\cite{utgoff1989incremental} & &\multicolumn{2}{c}{}\\
  \hline
  $\mathcal{X}-$   &  &  & Co-training\cite{blum1998combining} &\multicolumn{2}{c}{}\\
  &  &   &Heterogeneous transfer learning\cite{wang2011heterogeneous,zhu2011heterogeneous} & \multicolumn{2}{c}{Analogy learning~\cite{wang2011transfer}}\\
  \hline
\end{tabular}
}
\end{table*}

\subsection{Transfer learning methods}

We give a short description of the learning methods appearing in Table~\ref{tab:transf}. For each method, only the general idea is presented, and application of these methods to speech and language processing is left to the next sections.

\subsubsection{Model adaptation and incremental training}
\label{sec:tl:cat:adapt}

The simplest transfer learning is to adapt an existing model to meet the change of data distribution. Both the feature and label spaces are the same for the source and target domains, and the models are the same. There are various approaches for model adaptation. For example, the maximum \emph{a posterior} (MAP)~\cite{gauvain1994maximum} estimation and the maximum likelihood linear regression (MLLR) algorithm~\cite{leggetter1995maximum}. If the distribution changes gradually, then incremental or online learning is often used, e.g.~\cite{utgoff1989incremental,arandjelovic2006incremental,declercq2008online}.

Note that the adaptation can be either supervised or unsupervised. In the supervised learning, the data in the target domain are labelled, while in the unsupervised learning, no labels are available and they have to be generated by the model in the source domain before the adaptation can be performed. The latter case is often referred to as semi-supervised learning~\cite{zhu2005semi}. Note that semi-supervised learning is a general framework to deal with unlabelled data, and can be applied to any conditions where the label spaces are the same in the source and target domains. We will come back to this method in heterogeneous transfer learning that will be discussed shortly. Another approach to dealing with unlabelled data is to use them to derive new features (e.g., by linear projection) where the distributions of the data in the source domain and the target domain are close to each other. An interesting work towards this direction is the approach based on transfer component analysis (TCA)~\cite{pan2011domain}.

In another configuration, some unlabelled data are available but the distribution is different from that of the target domain. These data cannot be used for adaptation (either by semi-supervised learning or TCA) otherwise the model will be adapted to a biased condition.  However, it can be used to assist deriving more robust features. The idea is similar to TCA, but the unlabelled data are not used as supervision about the target domain, instead as an auxiliary information to derive more domain-independent features. This approach is often referred to as self-taught learning~\cite{raina2007self}, and it essentially holds the same idea as the more recent deep representation learning that will be discussed in Section~\ref{sec:tl:dp}.

\subsubsection{Heterogeneous transfer learning}

A more complex transfer learning scenario is to keep the labels and model unchanged, however the features are different in the source and target domains. The transfer learning in this scenario is often called heterogeneous transfer learning. The basic assumption for heterogeneous transfer learning is that some correspondence between the source and target domains exist, and this correspondence can be used to transfer knowledge in one domain to another. For example, speech and text are two domains, and there is clear correspondence between the two domains based on human concepts: no matter we speak or write `chicken', it is clear that we refer to the same bird that has wings but can not fly much.

The early research tried to define and utilize the correspondence between the instances of the source and target domains. For example, \cite{prettenhofer2011cross} employed an oracle word translator to define some pivot words that were used to establish the cross-domain correspondence by learning multiple linear classifiers that predict the `joint existence' of these words in the multi-domain data. In~\cite{dai2008translated} some instance-level co-occurrence data were used to estimate the correspondence in the form of joint or conditional probabilities; this correspondence was then used to improve the model in the target domain by risk-minimization inference. Asymmetric regularized cross-domain transformation was proposed in~\cite{kulis2011you}, which tries to learn a non-linear transform between the source and target domains by class-labeled instances from both source and target domains.  Although an instance does not necessarily possess features of both domains, the class labels offer the correspondence information.

More recent approaches prefer to finding common representations of the source and target domains, for example by matrix factorization~\cite{zhu2011heterogeneous}, RBM-based latent factor learning~\cite{wei2011heterogeneous}, or joint transfer optimization~\cite{shi2010transfer,wang2011heterogeneous,duan2012learning}. More recently, deep learning and heterogeneous transfer learning are combined where high-level features are derived by deep learning and inter-domain transforms are learned by transfer learning~\cite{zhou2014hybrid}.

We emphasize that most of the approaches discussed above assume that the label space does not change when transferring from the source domain to the target domain. A more ambitious task is to learn from very different tasks for which the label space is different from the target domain. For example, the task in the source domain is to classify document sentiment, while in the target domain the task is to classify image aesthetic value. This two tasks are fundamentally different, however some analogy does exist between them.
Learning correspondence between two independent but analogous domains is easy for humans~\cite{gentner1983structure,gentner1997reasoning,blitzer2006domain}, however it is very difficult for machines. There has been long-term interest in analogy learning among artificial intelligence researchers, e.g.,~\cite{carbonell1983learning,wang2011transfer}, though not too much achievement yet. Interestingly, the recent improvement in deep learning methods seems provide more hope in this direction, by a unified framework for representation learning and multitask learning. This will be discussed in Section~\ref{sec:tl:dp}.

\subsubsection{Multiview co-training}


A special case of heterogeneous transfer learning is the multi-view co-training, which assumes that each training instance involves features of both the source and target domains, but only the feature in the target domain is available at runtime. In this condition, heterogeneous transfer learning is not very effective since the training instances in the source domain are the same as the instances in the target domain and so does not provide much additional information, at least with supervised learning. However, the multi-view property of the training data indeed can be used to improve unsupervised learning with unlabelled data, by the approach called co-training~\cite{blum1998combining}. Specifically, co-training trains two separate models with features of the source and target domains respectively, and then generates labels for the unlabelled data using one model, which are in turn used to update the other model. This process iterates until convergence is obtained. It is well-known that co-training leads to better models than training with the feature of the target domain only.

\subsubsection{Model transfer}

If the feature and label spaces are the same however the models are different for the source and target domains, the knowledge learned by the source model can be transferred to the target model by model transfer. For example, in the source domain the model is a Gaussian mixture model (GMM), while in the target domain the model is a deep neural network (DNN). The transfer learning then exploits the GMM to initialize and boost the DNN. This is the general recipe in the modern DNN-based speech recognition system. Recently, this model transfer has gained much attention in the deep learning community. For example, it is possible to learn simple neural nets from a complex DNN model, or vice versa~\cite{hinton2014distilling,wang2015recurrent,tang2015knowledge}. Some interesting work in this direction will be presented in the next sections.

\subsubsection{Multitask learning}

In the case where the feature spaces of the source and target domains are the same but the task labels are significantly different, multitask learning is more applicable~\cite{caruana1997multitask,baxter2000model,guinney2011estimating}. The basic assumption of this learning approach is that the source and target tasks are closely related, either positively or negatively, so that learning for one task helps learning the other in the form of mutual regularization. Multitask learning is a general approach that can be applied to boost various types of models including kernel regression, k-nearest neighbour, and it can be even employed to learn `opposite' tasks simultaneously, e.g., text content classification and emotion detection~\cite{romera2012exploiting}.

A particular issue of multitask learning is how to evaluate the relevance of two tasks so that whether they can be learned together can be determined. Although there is not a simple solution yet, \cite{guinney2011estimating} indeed provided an interesting approach that estimates the relevance between tasks by evaluating the overlap of different tasks in the same semantic space.

\subsection{Transfer learning in deep learning era}
\label{sec:tl:dp}

Deep learning almost changed everything, including transfer learning. Because deep learning gains so much success in speech and language processing~\cite{dnn:hinton12:141,deng2014,he2014deep,hirschberg2015advances}, we put more emphasis on transfer learning methods based on deep models in this paper.
Roughly speaking, deep learning consists of various models that involve multi-level representations and the associated training/inference algorithms. Typical deep
models include deep belief networks (DBNs)~\cite{hinton2006fast}, deep Boltzmann machines (DBMs)~\cite{salakhutdinov2009deep}, deep auto encoders (DAEs)~\cite{bengio2007greedy,vincent2010stacked}, deep neural networks (DNNs)~\cite{dahl2012context,deng2014} and deep recurrent neural networks (RNNs)~\cite{graves2013speech}.

The success of deep models is largely attributed to their capability of learning multi-level representations (features), which simulates the processing pipeline of human brains where information is processed in a hierarchical way. The multi-level feature learning possesses several advantages. First, it can learn high-level features which are more robust against data variation than features at low-levels; second, it offers a hierarchal parameter sharing that holds great expressive power~\cite{bengio2011expressive}; third, the feature learning can be easily conducted without any labelled data and so is cheap; fourth, with a little supervised training (fine-tuning), the learned models can be well adapted to specific tasks~\cite{caruana1997multitask,collobert2008unified,deng2013recent}.

For these reasons, deep learning provides a graceful framework for transfer learning, which unifies almost all the approaches listed in Table~\ref{tab:transf} as \emph{representation learning}. The basic idea is to learn some high-level robust features that are shared by multiple features and multiple tasks, so that all the knowledge/model transfers are implemented as feature transfer. This approach was advocated in the NIPS95 workshop as a major research direction, but it was not such successful until deep learning became a main stream in machine learning and related fields~\cite{gutstein2010transfer,bengio2012deep,ngiam2011multimodal,Bengio-et-al-2015-Book}.

\begin{figure}[htb]
  \centering{
  \epsfig{file=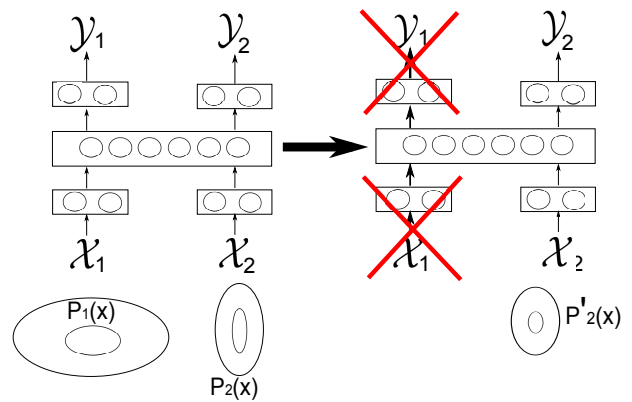, width=80mm}
  \caption{Transfer learning architecture with deep representation learning. $\mathcal{X}_1$ and $\mathcal{Y}_1$ are the feature and label spaces respectively for the learning task in the source domain, and $\mathcal{X}_2$ and $\mathcal{Y}_2$ are for the learning task in the target domain. At the runtime, only the target domain is concerned. }
  \label{fig:dp}
  }
\end{figure}

The transfer learning architecture based on deep representation learning is illustrated in Fig.\ref{fig:dp}. The left part of this figure is the joint training phase where heterogeneous input features are projected onto a common semantic space by different pre-processing networks, and the shared features involve rich explanatory factors that can be used to perform multiple tasks. The right part of the picture illustrates the adaptation phase, where some data $\mathcal{X}_2$ for the target task $\mathcal{Y}_2$ have been provided, either with or without labels, and the model is updated with the new data which follows a distribution $P'_2(x)$ that is different from the original distribution $P_2(x)$ in the joint training phase.

The framework in Fig.~\ref{fig:dp} is very flexible and covers almost all the methods in Table~\ref{tab:transf}. For example, without the adaptation phase, it is basically a multitask learning, while using multi-domain data also implements structural correspondence learning and latent representation learning. If the joint training phase involves only a single task, then the adaptation phase implements the conventional model adaptation. It should be highlighted that a particular advantage of the representation learning framework is that the feature extractor can be trained in an unsupervised way, e.g., by restricted Boltzmann machines (RBMs)~\cite{hinton2006reducing} or auto-associators~\cite{bengio2007greedy}, therefore little or no labelled data are required. According to~\cite{bengio2012deep}, as long as the distribution $P(X)$ is relevant to the class-conditional distribution $P(Y|X)$, the unsupervised learning can improve the target supervised learning, in terms of convergence speed, amount of labelled data required and model quality.

An early work based on deep representation learning is~\cite{glorot2011domain}, where the authors used unsupervised learning (denoising auto-encoders) to extract high-level features, and trained a sentiment analysis system in one domain (e.g., book review). They found that the system could be directly migrated to a new domain (e.g., DVD review) and achieved much better performance than competitive approaches including structural correspondence learning (SCL) and spectral feature alignment (SFA). This work demonstrated that high-level abstract features are highly domain-independent and could be easily transferred across domains, even without any adaptation. As another example, \cite{oquab2014learning} showed that CNN-based representations learned from a large image database imageNet were successfully applied to represent images in another database PASCAL VOC. A similar study was proposed recently in~\cite{zhang2015deep} where CNN features trained on multiple tasks were successfully applied to analyze biological images in multiple domains.

In another example called `one-short learning'~\cite{fei2006one}, high-level features trained on a large image database were found to be highly generalizable, and a very few labeled data could adapt models to recognize unseen objects by identifying the most relevant features. In a more striking configuration, the learning task can be specified as an input vector (task vector, e.g., a vector that represents a subregion of the data where the classification takes place) and fed into the deep nets together with the input data. The network then can learn the complex relationship between the data vector, the task vector, and the task labels. As long as the new task can be related to the tasks seen in the training phase (which can be obtained by a distributed task vector with which the relation between tasks can be estimated from the distance between task vectors), the new task can be well performed without any adaptation. This leads to the zero-data learning~\cite{larochelle2008zero} and zero-shot learning ~\cite{socher2013zero}.

\section{Transfer learning in speech processing}
\label{sec:speech}

Speech signals are pseduo-stationary and change vastly according to a large number of factors (language, gender, speaker, channel, environment, emotion, ...).  Dealing with these varieties is the core challenge of the speech processing research, and transfer learning is an important tool to solve the problem. It is not possible to cover all the researches in a short paper, so we select three most prominent fields where transfer learning has gained much success: transfer across languages, transfer across speakers, and transfer across models.


\subsection{Cross-lingual and multilingual transfer}

It is natural to believe that some common patterns are shared across languages. For example, many consonants and vowels are shared across languages, defined by some universal phone sets, e.g., IPA. This sharing among human languages have been utilized explicitly or implicitly to improve statistical strength in multilingual conditions, and has delivered better models than training with monolingual data, especially for low-resource languages. This advantage has been demonstrated in a multitude of research fields, though our review simply focuses on speech recognition and speech enhancement.

Early approaches to employing cross-lingual or multilingual resources is via some linguistic correspondence, e.g., by a universal phone set or a pair-wised phone mapping~\cite{schultz2001language,vu2011cross}. With the popularity of deep learning, the DNN-based multilingual approach in the form of representation learning gained much interest. The basic idea is that the features learned by DNN models tend to be language-independent at low layers and more language-dependent at high layers. Therefore multilingual data can be used to train a multilingual DNN where the low-level layers are shared by all languages, while the high-level layers are language specific. This is fully consistent with the representation learning framework shown in Fig.~\ref{fig:dp}, where $\mathcal{Y}_1$ and $\mathcal{Y}_2$ represent two languages. By this sharing diagram, the features can be better learned with multilingual data, and for each language, training only the language-specific part is much easier than training the entire network.

The initial investigation was proposed in~\cite{swietojanski2012unsupervised}, where multilingual data were used to initialize the DNN model for the target language. Interesting improvement was reported and this approach was quickly followed by researchers, with both the DNN-HMM hybrid setting and the tandem setting.

With the hybrid setting, DNNs are used to replace the conventional GMMs to estimate the likelihood of HMM states. In the multilingual scenario, the hidden layers of the DNN structure are shared across languages and each language holds its own output layer~\cite{huang2013cross,heigold2013multilingual,ghoshal2013multilingual}. The training process then learns a shared feature extractor as well as language-dependent classifiers. This approach was proposed independently by three research groups  in 2013, and tested on three different databases: English and Mandarin data~\cite{huang2013cross}, eleven Romance languages~\cite{heigold2013multilingual} and the global phone dataset with 19 languages~\cite{ghoshal2013multilingual}. A simple extension of the above setting is to involve multiple layers in the language-specific part, or simply use different classifiers (the default is software regression), although the latter is much similar to the tandem approach discussed below.

With the tandem setting, DNNs are used as feature exactors, based on which posterior features or bottleneck features are obtained and are used to train conventional GMM-HMM systems. In~\cite{vesely2012language,thomas2013deep}, the same DNN structure as in the hybrid setting was used to train a multilingual DNN, however the model was used to produce features (from the last hidden layer) instead of state likelihood. It was showed that the features generated by multilingual DNNs are rather language-independent and can be used directly for new languages.  With limited adaptation data in the target language, additional performance could be obtained. The same approach was proposed in~\cite{tuske2013investigation}, though the features were read from a hidden layer in the middle layer (the bottle net layer with less neurons than other layers) instead of the last hidden layer. The features produced in this way are often referred to as bottleneck (BN) features. Combing with a universal phone set, the language-independent BN features can be used to train models for languages even without any labelled data~\cite{knill2014language}.

The hybrid setting and tandem setting can be combined. For example, in~\cite{bell2013multi}, the BN feature was first derived from a multilingual DNN, and then it  was combined with the original feature to train a hybrid system. A similar approach was proposed in~\cite{gehring2013dnn}, where the BN feature extractor for each language was regarded as a module, and another DNN combined the BN features from the modules of multiple languages to construct the hybrid system.

The multilingual DNN approach described above belongs to multitask learning which can be extended to more general settings. For example, in~\cite{chen2014joint} phone recognition and grapheme recognition were treated as two different tasks to supervise the DNN acoustic model training. They tested on three low-resource south African languages and showed that the mutitask training indeed improved performance. They also compared the multitask training with the conditional training where the grapheme recognition provided additional input for the phone recognition, instead of co-supervision.

In a slightly different configuration, we reported a multitask learning which learns speech content and speaker accent together~\cite{tang152}.  In this approach,
a pronunciation vector that represents the accent of a speaker is generated by either an i-vector system~\cite{dehak2011front} or a DNN system~\cite{Ehsan14}. This pronunciation vector can be integrated in the input or hidden layers as additional features (the conditional learning), or used as an auxiliary output of a hidden layer (the multitask learning).
In the latter setting, the pronunciation vector plays the role of a regularization to help learn better representations that can disentangle the underlying factors of the speech signal. We tested the method in an English ASR task where the speech data are in multiple accents (British and Chinese). We found that both the two approach could improve performance for utterances in both British and Chinese accents. An advantage with the second setting, however, is that the pronunciation vector is required only at the training phase. This is actually a heterogeneous multitask learning that has been proposed for a long time~\cite{caruana1997multitask} but has not been studied much in speech processing.

Besides speech recognition, cross-lingual and multilingual transfer were also proposed for speech enhancement. The assumption is that the noise and reverberation that need to be removed are largely language-independent, and therefore an enhancement model trained with the data in one language can be applied directly to other languages. For example, in~\cite{xu2014cross}, an DNN architecture trained in English data was demonstrated to be highly effective for enhancing speech signals in Chinese, by re-using the first several layers which were assumed to be language-independent. Another study published recently from our group demonstrated that a DNN structure can be used to remove music from speech in multilingual conditions~\cite{zhao15}.


\subsection{Speaker adaptation}

Speaker adaptation is another domain in which transfer learning has gained brilliant success. In the paradigm of parametric statistic models (e.g., Gaussian models or Gaussian mixture models), maximum a posterior (MAP) estimation~\cite{gauvain1994maximum} and maximum likelihood linear regression (MLLR)~\cite{leggetter1995maximum} are two most successful methods to adapt a general model to a specific speaker.  These methods are still the research focus of some authors, e.g.~\cite{seltzer2011separating,povey2012basis,miao2013learning}. A short survey for these early-stage techniques can be found in~\cite{shinoda2011speaker}.

In the deep learning era, DNN models are widely used nearly everywhere. However, adapting neural network, particular a deep one, is not simple, because DNN is a highly compact distributed model.  It is not easy to learn a simple form (with limited amount of parameters) such as MLLR to update all parameters of the network. However, recent research shows that with some particular constrains on the adaptation structure, speaker adaptation is possible for DNN models.

An early study reported in~\cite{abdel2013fast} introduced a user vector (user code) to represent a speaker, and the vector was augmented to the input and hidden layers. The learning then trained the network and the speaker code simultaneously. To adapt to a new speaker, the network was fixed while the speaker vector was inferred by the conventional back-propagation algorithm~\cite{abdel2013rapid}. This approach was extended in~\cite{xue2014direct} by involving a transform matrix before the speaker vector was augmented to the input and hidden layers, possibly in the form of low-rank matrices.

In a similar setting, the speaker code can be replaced by a more general speaker vector produced by exotic models, e.g., the famous i-vector~\cite{dehak2011front}. Different from the speaker code approach, these speaker vectors do not need to be adapted (although it is possible)~\cite{saon2013speaker,karanasou2014adaptation,senior2014improving,gupta2014vector}. An advantage of using exotic speaker vectors is that the speaker vector model can be trained with a large unlabelled database in an unsupervised fashion. A disadvantage is that no phone information is considered when deriving the vectors, at least it is case with the i-vector model. A careful analysis for the i-vector augmentation was conducted in~\cite{rouvier2014speaker}, which showed that i-vectors not only compensate for speaker variance but also acoustic variance.

In contrast to involving an speaker vector, the second approach to speaker adaptation for DNN models is to update the DNN model directly, with some constraints on which components of the DNN should be adapted. For example, the adaptation can be conducted on the input layer~\cite{neto1995speaker,yao2012adaptation}, the activations of  hidden layers~\cite{gemello2007linear,siniscalchi2013hermitian,swietojanski2014learning}, or the output layer~\cite{yao2012adaptation}. Some comparison for adaptation on different components can be found in~\cite{li2010comparison,liao2013speaker}. In order to constrain the adaptation more aggressively, \cite{xue2014singular,xue2014speaker} studied a singular value decomposition (SVD) approach which decomposes a weight matrix as production of low rank matrices, and only the singular values are updated for each speaker. Another constraint for speaker adaptation is based on a prior distribution over the output of the adapted network, which is imposed by the output of the speaker-independent DNN, in the form of KL-divergence~\cite{yu2013kl}.

Another interesting approach to speaker adaptation for DNN models is to apply transfer learning to project features to a canonical speaker-independent space where a model can be well trained. For example, the famous constrained MLLR (CMLLR) in the HMM-GMM architecture~\cite{leggetter1995maximum}. Recently, an auto-encoder trained with speaker vectors (obtained from a regression-based speaker transform) was used to produce speaker-independent BN features~\cite{tang2014deep}. A similar approach was studied in~\cite{miao2014towards}, though an i-vector was used as the speaker representation.

Most of the above researches are based on the DNN structure. Recent research shows that RNNs can be adapted in a similar way. For example, \cite{miao2015speaker} reported an extensive study on speaker adaptation for LSTMs. It was found that LSTMs can be effectively adapted by using speaker-adaptive (SA) front-end (e.g., a speaker-aware DNN projection~\cite{miao2014towards}), or by inserting speaker-dependent layers.

It should be noted that DNN itself possesses great advantage of learning multiple conditions. Therefore, DNN models trained with a large amount of data of multiple speakers can deal with speaker variation pretty well. This conjecture was demonstrated by~\cite{liao2013speaker}, which showed that the adaptation methods provide some improvement if the network is small and the amount of training data is medium, however for a large network trained with a large mount of data, the improvement is insignificant.

The techniques discussed above are mostly applied to speech recognition, however they can be easily migrated to other applications. For example in HMM-based speech synthesis, model adaptation based on MAP and MLLR has been widely used to produce specific voice, e.g.,~\cite{tamura2001adaptation,wu2009state,yamagishi2007average,yamagishi2009analysis}.
Particularly, speaker adaptation is often coupled with language adaptation to obtain multilingual synthesis, e.g., by state mapping~\cite{wu2009state,liang2010comparison,gibson2010unsupervised}. For DNN-based speech synthesis~\cite{ling2013modeling,zen2014deep,hashimoto2015effect}, it is relatively new and the adaptation methods have not been extensively studied, except a few exceptions~\cite{potard2015preliminary,wu2015}.


\subsection{Model transfer}

A recent progress in transfer learning is to learn a new model (child model) from an existing model (teacher model), which is known as model transfer. This was mentioned in the seminal paper of multitask learning~\cite{caruana1997multitask} and was recently rediscovered by several researchers in the context of deep learning~\cite{ba2014deep,hinton2014distilling,li2014learning}. The initial idea was that the teacher model learns rich knowledge from the training data and this knowledge can be used to guide the training of child models which are simple and hence unable to learn many details without the teacher's guide. To distill the knowledge from the teacher model, the logit matching approach proposed by Ba~\cite{ba2014deep} teaches the child model by encouraging its logits (activations before softmax) close to those generated by the teacher model in terms of square error, and the dark knowledge distiller model proposed by Hinton~\cite{hinton2014distilling} encourages the output of the child model close to those of the teacher model in terms of cross entropy.

This approach has been applied to learn simple models from complex models so that the simple model can approach the performance of the complex model. For example, \cite{li2014learning} utilized the output of a complex DNN as regularization to learn a small DNN that is suitable for speech recognition on mobile devices. \cite{chan2015transferring} used a complex RNN to train a DNN. Recently, a new architecture called FitNet was proposed~\cite{romero2014fitnets}. Instead of regularizing the output, FitNet regularizes hidden units so that knowledge learned by the intermediate representations can be transferred to the target model, which is suitable for training a model whose label space is different from that of the teacher model. This work was further extended in~\cite{long2015learning}, where multiple hidden layers were regularized by the teacher model. Another example is to transfer heterogeneous models. For instance, in~\cite{lu2015unsupervised}, unsupervised learning models (PCA and ICA) were used to model the outputs of a DNN model. This in fact treats the DNN output as an intermediate feature, and uses the feature for general tasks, e.g., classifying instances from novel classes.

Our recent work~\cite{wang2015recurrent} showed that this model transfer can not only learn simple models from complex models, but also the reverse: a weak model can be used to teach a stronger model. In our work~\cite{wang2015recurrent}, a DNN model was used to train a powerful complex RNN. We found that by the model transfer learning, RNNs can be learned pretty well with the regularization of a DNN model, though the teacher model is weaker than the target one. In a related work~\cite{tang2015knowledge}, we found that the model transfer learning can be used as a new pre-training approach, and it even works in some scenarios where the RBM pre-training and layer-wised discriminative pre-training do not work. Additionally, combining the RMB-based pre-training and the model transfer pre-training can offer additional gains, at least in our experimental setting where the training data is not very abundant.





\section{Transfer learning in language processing}
\label{sec:lang}

As in speech processing, the basic assumption of transfer learning for language processing is also intuitive: all human languages share some common semantic structures (e.g., concepts and syntactic rules). Following this idea, the simple way of transfer learning in multilingual or multi-domain scenarios is to construct some cross-lingual/cross-domain correspondence so that knowledge learned in one language/domain can be transferred and reused in another language/domain. For example, a bi-lingual lexicon can be used to provide instance-level correspondence so that syntactic knowledge learned in one language can be used to improve the syntactic learning in the second language~\cite{durrett2012syntactic}. Another approach that gained more attention recently is to learn a common latent space that are shared by different languages or domains, so that knowledge can be aggregated, leading to improved statistic strength for probabilistic modeling in each single language or domain.

Once again, transfer learning is such a broad research field and the research of language processing is even more broad itself, which makes a detailed review for all the research fields impossible in such a short paper. We will focus on two particular fields: cross-lingual learning and cross-domain learning, particularly for the document classification task.

\subsection{Cross-lingual and multilingual transfer learning}

A straightforward way to transfer knowledge between languages is to translate words from one language to another by a bi-lingual lexicon. For example, this approach was used in~\cite{shi2010cross} to translate a maximum entropy (ME) classifier trained in English data to a classifier used for classifying Chinese documents. In another work from our group, we have applied this approach successfully to train multilingual language models, where some foreign words need to be addressed~\cite{ma15}. Word-by-word translation, however, is obviously not ideal since no syntactic constraints in different languages are considered. A more complicated approach is to translate the whole sentence by machine translation~\cite{koehn2009statistical}, so that any labelling or classification tasks in one language can be conducted with models trained in another language.


A more recent approach to multilingual learning is to learn some common latent structures/representations based on multilingual data. For example, the multilingual LDA model proposed in~\cite{de2011knowledge} assumes a common latent topic space, so that words from multiple languages can share the same topics. This is similar to the RMB-based heterogeneous factor learning~\cite{wei2011heterogeneous}: both are based on unsupervised learning with weak supervision, i.e., no word alignment is required.

A similar approach proposed in~\cite{tackstrom2012cross} learns multilingual word clusters, where a cluster may involve words from different languages. This was achieved by means of a probabilistic model over large amounts of monolingual data in two languages, coupled with parallel data through which cross-lingual correspondence was obtained. Applying to the NER task, it was found that up to 26\% performance improvement was observed with the multi-lingual model. This work was extend in~\cite{tackstrom2012nudging} where cross-lingual clusters were used to `directly' transfer an NER model trained in the source language to the target language.

Another approach to constructing common latent space is by linear projection instead of statistical models. For example, in the heterogeneous feature augmentation (HFA) approach proposed in~\cite{duan2012learning}, two linear projections are learned to project features in different languages to a common space. In their study, these projections were used to produce additional features that were augmented to the original features to train the model in the target language. An interesting part of their approach is to train the supervision model (e.g., SVM) in the source and target languages \emph{simultaneously}. This leads to a joint optimization for the common space projections as well as the classifiers. The approach was tested on a text classification task with the Reuters multilingual database and obtained good performance. In another work~\cite{prettenhofer2011cross}, a linear projection was learned by optimizing a set of multi-lingual classifier, each of which predicted the existence of the words of a bi-lingual word-pair. The approach was tested on cross-lingual topic discovery and sentiment classification.

Recently, word embedding becomes a hot topic~\cite{bengio2006neural,mnih2008scalable,turian2010word,collobert2011natural,mikolov2013efficient}. Intuitively, word embedding represents each word as a low-dimensional dense vector (word vector) with the constraint that relevant words are located more closely than irrelevant words. This embedding enables semantic computing over words, and provides new ways for mulitilingual learning: if word vectors can be trained in a multilingual fashion, regressors/classifiers trained on these vectors naturally apply to multiple languages.

A simple approach is to map word vectors trained in individual languages to a single space. For example, in~\cite{mikolov2013exploiting}, it was found that a linear transform can project word vectors trained in one languages to word vectors in another language so that relevant words are put closely, in spite of their languages. This projection can be learned simply by some pivot word pairs from the two languages. We extended this work in~\cite{xing15} by modeling the transfer as an orthogonal transform. A more systematic approach was proposed by~\cite{faruqui2014improving}, where different languages were projected to the same space by different projections, and the projections were determined by maximizing the canonical correlation of the corresponding words in the projected space. This approach requires one-to-one word correspondence, which was obtained by aligned parallel data.

A potential problem of the above approaches is that the word vectors and projections are learned separately. The approach proposed in~\cite{klementiev2012inducing} does not learn any projection, instead the bi-lingual correspondence was taken into account in the embedding process. This work was based on the neural LM model~\cite{bengio2006neural} and changed the objective function by involving an extra term that encourages relevant words in different languages located together. The relevance of words in different languages was derived from aligned parallel data.

In another work~\cite{hermann2014multilingual}, the relevance constraint was employed at the sentence level. Word vectors were aggregated into a sentence embedding, and relevant sentences were embedded closely. This approach does not require word alignment and so can be easily implemented. Additionally, this approach can be simply extended to document level models, for which only document pairs are required, without any sentence-level alignment. This approach was tested on a multilingual classification task.

A similar work was proposed by~\cite{gouws2014bilbowa}. As in~\cite{hermann2014multilingual}, only sentence pairs are required in the learning; the difference is that the embedding leveraged both monolingual data and bi-lingual data, and employd noise-contrastive training to improve efficiency. Good performance was obtained in both cross-lingual document classification and word-level translation.


An interesting research that involves much ingredient of deep learning was proposed by~\cite{zhou2014hybrid}. The basic idea is to learn high-level document features individually in each language by unsupervised learning (i.e., mSDA in that paper), and then learn the correspondence (transform) using parallel data. The raw and high-level features can be combined to train the classifier in one language, and documents in another language can be transferred to the rich language and are classified there. The idea of applying unsupervised learning to learn high-level features is prominent, which may help remove noises in the raw data thus leading to more reliable transform estimation. The approach was tested on several multilingual sentiment classification tasks where the raw document feature was TF-IDF and the high-level features were learned by mSDA. Good performance was reported.





\subsection{Cross-domain transfer learning}

Cross-domain transfer learning has two different meaning: when the domain refers to applications, then the difference is in the data distribution; when it refers to features, then the difference is in feature types or modalities, e.g., audio feature or image feature.  We focus on the feature domain transfer, which is relatively new and invokes much interest recently. With the simplest approach, multi-modal features can be combined either at the feature level or the score level. For example on the semantic relatedness task, \cite{bruni2012distributional} concatenated visual and textual features to train multi-stream systems;  in~\cite{leong2011going},  the scores predicted by multiple models based on different features are combined. A more complex setting involves transferring knowledge between models built with heterogeneous features. Note that some authors regard different languages as different domains, e.g.,~\cite{zhou2014hybrid}. However, we focus on transfer learning between different feature modalities.

An example is the work proposed in~\cite{dai2008translated}, where the authors used co-occurrence data to estimate the correspondence between different features,
i.e., image and text. The estimated correspondence was then used to assist the classification task in the target domain, by transferring the target features to the source domain where a good classification model had been constructed. The authors formulated this transfer process using a Markov chain and risk minimization inference. The method was tested on a text-aided image classification task and achieved significant performance improvement.

The common latent space approach was studied in~\cite{socher2010connecting}, with the task of image segmentation and labelling. The model was based on kernelized canonical correlation analysis which finds a mapping between visual and textual representations by projecting them into a latent semantic space.

Deep learning provides an elegant way for cross-domain transfer learning, with its great power in learning high-level representations shared by multiple modalities~\cite{ngiam2011multimodal}. For example, in~\cite{socher2013zero,frome2013devise}, images and words are embedded in the same low-dimensional space via neural networks, by which image classification can be improved by the word embedding, even for classes without any image training data. \cite{kiros2014multimodal} proposed a multi-modal neural language modeling approach with which the history and prediction can be both text and images, so that the prediction between multiple modalities becomes possible. In~\cite{socher2014grounded}, an RNN structure based on dependency-tree was proposed to embed textual sentences into compositional vectors, which were then projected together with image representations to a common space. Within this common space, multi-modal retrieval and annotation can be easily conducted. The same idea was proposed by~\cite{srivastava2012multimodal}, though deep Boltzmann machines were used instead of DNNs to infer the common latent space.


\subsection{Model transfer}

Model transfer, which aims to learn one model from another, has not yet been extensively studied in language processing. A recent work~\cite{mou2015distilling} studied a knowledge distilling approach on the sentiment classification task. The original classifier was a large neural net with large word vectors as input, and a small network was learned in two ways: either using the output of the large network as supervision or directly transferring large word vectors to smaller ones.

In a recent study~\cite{zhang2015learning}, we show that it is possible to learn a neural model using supervision from a Bayesian model. Specifically, we tried to learn a document vector from the raw TF input using a neural net, supervised by the vector representation produced by latent Dirichlet allocation (LDA). Our experimental results showed that with a two-layer neural network, it is possible to learn document vectors that are quite similar to the ones produced by LDA, while the inference is hundreds of times faster.

\section{Perspective and conclusions}
\label{sec:con}

We gave a very brief review of transfer learning, and introduced some applications of this approach in speech and language processing. Due to the broad landscape of this research and the limited knowledge of the authors, only very limited areas were touched. Also, many important contributions in the `history' had to be omitted, for the sake of emphasis on more recent directions in the past few years, especially deep learning. Even with such a limited review, we can still clearly see the important role that transfer learning plays and how fast it has evolved recently. For speech and language processing, transfer learning is essentially important as both speech and language are diverse, imbalance, dynamic and inter-linked, which makes transfer learning inevitable.

Transfer learning can be conducted in very different manners. It can be conducted as a shared learning that learns various domains and tasks together, or as a tandem learning which learns a model in one domain/task and migrates the model to another domain/task. It can be conducted with a supervised way where labeled data are used to refine the classifier, or an unsupervised way where numerous unlabelled data are used to learn better representations. It can be used to transfer instances, representations, structures and models. It can transfer between different distributions, different features and different tasks.

Go back to the NIPS 95 workshop, where some questions were raised by the famous researchers at that time. Two decades later, we can answer some of the questions, while other remains mystery:

\begin{itemize}

\item {\it What do we mean by related tasks and how can we identify them? } It is still difficult to measure relatedness, particularly with the complex configurations of transfer learning. However, we do know some possible metrics, e.g., the relatedness between marginal and conditional distributions~\cite{bengio2012deep} in unsupervised feature learning, or representation overlap in model adaptation~\cite{guinney2011estimating}. Particularly, we now know that even two tasks are intuitively unrelated (e.g., speech recognition and speaker recognition), transfer learning still works by utilizing the fact that the tasks are unrelated~\cite{romera2012exploiting}.

\item {\it How do we predict when transfer will help (or hurt)?} Again, it is not easy to find a complete solution. However some approaches indeed can alleviate negative transfer, e.g., \cite{long2014transfer,guinney2011estimating}. With deep learning, the risk of negative transfer seems substantially reduced. For example, any data in related domains can be used to assist learning abstract features, even they are sampled from a distribution different from the target domain~\cite{raina2007self}. This is not the case twenty years ago.

\item {\it What are the benefits: speed, generalization, intelligibility,...?} Seems all of these can be improved by transfer learning.

\item {\it What should be transferred: internal representations, parameter settings, features,...?} We now know all these components can be transferred.

\item {\it How should it be transferred: weight initialization, biasing the error metric,...?} All these methods can be used, although it seems that the regularization view is more attractive and it is related to modifying the objective function.

\item {\it How do we look inside to see what has been transferred?} This question is more related to model adaptation and the answer is model-dependent. For example with a DNN model which is highly compact, it is not simple to investigate which part of the model has been changed after adaptation.

\end{itemize}

Transfer learning has been widely studied in speech and language processing, particularly for model adaptation. Recent advance in multilingual learning and heterogeneous feature transform demonstrates the power of transfer learning in a more clear way. Nevertheless, compared to the very diverse methods studied in the machine learning community, application of transfer learning in speech and language research is still very limited. There are many questions remain unanswered, for example: can we learn common representations for both speech, language and speaker recognition? Can we learn acoustic models for voiced speech and whistle speech together? How about sign language? How to use large volume of unlabeled video data to regularize speech models? How pronunciation models can be used to regularize NLP tasks? How to involve heterogeneous resources including audio, visual, language to solve the most challenging tasks in the respective research fields? How to utilize the large amount of unlabeled data more efficiently in the big-data era? To solve these problems, we believe collaboration among researchers who have been used to work independently in their own areas is mostly required.

\section*{Acknowledgement}

This research was supported by the National Science Foundation of China (NSFC) under the project No. 61271389 and No. 61371136, and the National Basic Research Program (973 Program) of China under Grant No. 2013CB329302. It was also supported by the MESTDC PhD Foundation Project No.20130002120011, as well as Sinovoice and Huilan Ltd.
Thanks to Zhiyuan Tang for the careful reference collection.

\bibliographystyle{IEEEtran}
\small{
\bibliography{reference}
}

\end{document}